\documentclass[letterpaper, 10 pt, conference]{ieeeconf}  

\IEEEoverridecommandlockouts                              

\usepackage{cite}
\usepackage{amsmath,amssymb,amsfonts}
\usepackage{algorithmic}
\usepackage{graphicx}
\usepackage{todonotes}
\usepackage{textcomp}
\usepackage{xcolor}
\usepackage{booktabs}
\usepackage{multirow}
\usepackage{float}
\usepackage{xspace}
\begin{document}
\raggedbottom
\setlength{\textfloatsep}{8pt}
\setlength{\floatsep}{8pt}
\setlength{\intextsep}{8pt}

\title{\LARGE \bf{
FOCUS: Fine-Grained Open-Vocabulary Change Detection for Uncertainty-Aware Semi-Static Scenes
}}

\author{Can Xu$^{1}$, Mingfeng Yuan$^{1}$, Mahan Mohammadi$^{1}$, and Steven L. Waslander$^{1}$
\thanks{$^{1}$All authors are with the University of Toronto Institute for Aerospace Studies and the University of Toronto Robotics Institute, Toronto, ON, Canada.
{\tt\small \{first.last\}@robotics.utias.utoronto.ca}}}

\maketitle

\begin{abstract}
Autonomous robots operating over long periods must keep their environmental memory up to date as the world changes between visits. In semi-static environments, an object may be replaced in place by a different but geometrically and semantically similar instance, making the identity change difficult to detect from geometry or coarse semantics alone.
We present \textbf{\emph{FOCUS}}, an uncertainty-aware framework for object-level change detection and map maintenance. We formulate semi-static memory maintenance as probabilistic inference that fuses geometric likelihood with appearance likelihood rendered from a 3D Gaussian map. A recursive three-state estimator maintains whether each mapped object is \textsc{persisted}, \textsc{replaced}, or \textsc{removed}. To account for imperfect 3DGS rendering, we model the rendered appearance evidence probabilistically rather than using it as a direct change score, with the model parameters automatically calibrated from a change-free replay of the initial mapping session.
We evaluate our method on a new Isaac Sim warehouse benchmark with ambiguous in-place replacements and on the real-world TorWIC dataset. It improves object-level replacement F1 from 0.20 to 0.84 over a probabilistic baseline and transfers to real-world data without manual parameter retuning.
\end{abstract}

\section{Introduction}
The rapid advance of foundation models in recent years has greatly expanded robots’ ability to recognize unseen objects and understand open-ended environments. This progress is pushing robotics beyond short-horizon perception and reactive behavior toward long-term autonomy, where robots must operate reliably over extended periods as the world evolves. A key step toward this goal is persistent embodied intelligence: the ability to maintain an up-to-date memory of the environment, understand how it changes, and use that knowledge to adapt future decisions. Such capability is essential for warehouse inventory auditing, long-term crop monitoring, construction-site inspection, and household service robots, where objects may be added, removed, moved, replaced, or undergo meaningful changes in appearance or state over time. These settings are naturally \emph{semi-static}: most of the environment remains stable, while a small but important subset undergoes persistent change. Reliable operation in such environments therefore requires robots to distinguish genuine changes from viewpoint variation, sensing noise, transient observations, and subtle differences between visually similar objects. 

Recent open-vocabulary scene representations and scene graphs have enabled robots to build increasingly rich object-centric memories of their environments~\cite{yuan2026star,gorlo2026daaam}.

Existing semi-static mapping methods have begun to address this problem through geometric and probabilistic consistency estimation~\cite{qian2022pocd,schmid2024khronos}. \textbf{However, these methods are primarily designed to detect coarse scene changes rather than fine-grained changes in object identity or state.} POV-SLAM~\cite{qian2023pov}, for example, probabilistically estimates whether mapped objects remain consistent with current observations, but assumes prior semantic knowledge of mapped objects and models object state largely through geometric consistency. Khronos~\cite{schmid2024khronos} similarly reasons about long-term scene evolution using geometric visibility and free-space evidence to verify whether previously observed structures remain present. These approaches are effective when changes produce a clear geometric signature, such as object addition, removal, or displacement. They become much more ambiguous under \emph{in-place replacement}, where an object is replaced by another instance with nearly identical geometry and coarse semantics. For example, two cardboard boxes may occupy the same volume and share the same category while differing in a fragile label, plastic wrapping, or an open/closed state. Such differences may generate little geometric residual, yet can be operationally important for applications such as warehouse inventory reconciliation or long-term crop-condition monitoring.

Appearance can be used to augment geometric change detection, but appearance evidence is itself uncertain. Dense visual features can distinguish objects that share similar geometry and semantics, and recent methods exploit image descriptors such as DINO-based features for association across observations~\cite{oh2026oasismap}. However, pairwise matching usually requires sufficiently overlapping views, which may not exist across long-term revisits. 3D Gaussian Splatting (3DGS) provides an attractive alternative by storing a view-conditioned appearance model that can render the expected appearance of a mapped object from the current camera pose~\cite{ge2025dynamicgsg,deng2025omnimap}. This enables direct comparison between the current observation and the expected appearance without requiring a co-visible reference image. However, rendered appearance can be unreliable due to limited object visibility, incomplete reconstruction, floaters, pose errors, and illumination changes. A single-frame appearance residual is therefore insufficient for reliable change detection. Simply accumulating more observations is also problematic: along patrol-style trajectories, robots often observe an object from highly similar viewpoints over consecutive frames, causing correlated evidence to be repeatedly counted and resulting in overconfident decisions.

\textbf{We therefore formulate semi-static memory maintenance as uncertainty-aware probabilistic inference over object state.} For each mapped object, our system maintains a recursive three-state belief over \textsc{persisted}, \textsc{replaced}, and \textsc{removed}. Geometry and appearance provide complementary evidence: geometric cues are effective for detecting disappearance and newly occupied space, while rendered appearance provides the fine-grained signal needed to distinguish a persisted object from an in-place replacement. To make this appearance evidence reliable, we maintain an object-specific appearance baseline together with its uncertainty, compensate for frame-level appearance shifts, suppress unstable rendered evidence, and reduce the influence of temporally correlated observations. The resulting geometry and appearance likelihoods are fused in a recursive Bayesian estimator whose parameters are calibrated from a change-free replay of the reference session. In this way, our framework shifts semi-static memory maintenance from asking only \emph{``has the mapped geometry changed?''} toward the finer question \emph{``is this still the same object?''}

We summarize our contributions as follows:
\begin{enumerate}
    \item We introduce an uncertainty-aware 3DGS appearance model for fine-grained object change detection, accounting for appearance variation and unreliable rendered evidence.

    \item We formulate object-level memory maintenance as a three-state recursive Bayesian estimator over \textsc{persisted}, \textsc{replaced}, and \textsc{removed}, fusing complementary geometric and appearance evidence while accounting for unstable and correlated observations.

    \item We introduce a multi-session warehouse benchmark for fine-grained semi-static changes, with challenging in-place replacements between geometrically and semantically similar objects and a complete set of object-level change annotations.
\end{enumerate}

\section{Related Work}
\subsection{Long-Term Object-Level Map Maintenance}
Long-term map maintenance has traditionally relied on geometric inconsistency across revisits. TSDF- and SDF-based methods detect persistent changes from reconstruction differences~\cite{fehr2017tsdf,fu2022planesdf}, while Khronos~\cite{schmid2024khronos} uses geometric and free-space evidence to infer long-term object persistence. Such methods are effective for additions, removals, and displacement, but can be ambiguous for in-place replacement between geometrically similar objects. To provide additional cues beyond geometry, recent object-centric maps incorporate semantic features and language descriptions~\cite{conceptgraphs,deng2025omnimap,yuan2026star,gorlo2026daaam}. However, coarse semantics may still be insufficient to distinguish visually similar instances. We instead target fine-grained object identity changes by combining geometry with uncertainty-aware, view-conditioned appearance evidence.

\subsection{Appearance-Based Scene Change Detection}
Appearance provides complementary evidence for cross-session change detection. Recent 3DGS-based methods exploit novel-view rendering to compare a reference scene with current observations. 3DGS-CD~\cite{lu2025_3dgscd} detects 2D object-level changes from rendered references and fuses them across views to recover 3D change masks and object transformations, while Galappaththige et al.~\cite{galappaththige2026changes} extend this setting to online scene change detection with multi-view fusion and change-guided 3DGS updates. DynamicGSG~\cite{ge2025dynamicgsg} is particularly related to our setting: it renders visible mapped objects, compares them with current observations using SSIM, and removes objects whose similarity falls below a fixed threshold, treating them as moved or substituted. In contrast, we treat rendered appearance as an uncertain identity observation rather than a hard change decision, and recursively fuse it with geometric evidence to distinguish \textsc{persisted}, \textsc{replaced}, and \textsc{removed} objects.

\subsection{Probabilistic Object-State Estimation}
Change evidence from individual observations can be unreliable due to sensor noise, localization error, and imperfect segmentation. Persistence filtering therefore accumulates evidence over time rather than making single-frame decisions. Hashemifar and Dantu~\cite{hashemifar2020persistence} incorporate feature-persistence reasoning into visual SLAM, while POCD~\cite{qian2022pocd} recursively estimates object stationarity from geometric and semantic evidence. Bogenberger et al.~\cite{bogenberger2026glasses}, referred to as Perceive-Semantix, incorporate geometric and semantic similarity for persistent object-instance tracking. Building on this probabilistic formulation, we explicitly distinguish \textsc{persisted}, \textsc{replaced}, and \textsc{removed} states and recursively fuse geometry with uncertainty-aware appearance evidence.

\subsection{Datasets for Semi-Static Change Detection}
Several datasets support evaluation in changing indoor environments. TorWIC~\cite{qian2022pocd} provides multi-session warehouse sequences with object additions, removals, translations, and rotations for map-maintenance evaluation. 3RScan~\cite{wald2019rio} contains repeated scans of indoor scenes with dense instance annotations, cross-session correspondences, and object transformations, primarily targeting object re-localization.

However, these datasets do not directly target fine-grained object-state estimation under ambiguous in-place replacement. Our benchmark therefore introduces geometrically and semantically similar replacements with per-object change states and globally consistent instance identities across sessions.

\section{System Description}
\subsection{Problem Formulation and Assumptions}
\label{sec:problem}

In this work, we consider object-level change detection and semantic map maintenance across multiple traversals of a semi-static environment. Specifically, the environment remains static during each visit, and object changes only occur when the robot is away. We construct an object-level reference map during the first traversal, capturing scene geometry, semantics, object identity, and appearance information. During each revisit, we detect newly appeared objects and recursively update the state belief for each mapped object to determine whether its state has changed, updating the reference map accordingly. We define three states for each object, $S_i \in \{\textit{pr} = \textsc{persisted}, \textit{rp} = \textsc{replaced}, \textit{rm} = \textsc{removed}\}$. \textsc{Persisted} denotes that the mapped object remains unchanged. \textsc{Replaced} denotes the mapped object has been removed and a new object occupies the same space with similar geometry. \textsc{removed} denotes the object no longer exists and its space is no longer occupied. Newly appeared objects do not belong to this state space; the system detects them using the reference-map construction pipeline. We do not model object motion: when an object is displaced, we treat it as a disappearance at the original pose and, following the rule above, as a newly appeared object at the new location. We leave for future work the simple extension to associate new objects with previously removed ones to handle displacement explicitly.

For each mapped object we maintain a belief over its state given the observations up to time $t$,
\begin{equation}
\begin{array}{l}
    b_i(S) = p(S \mid z_{1:t}), \\ \qquad S \in \{\textsc{persisted}, \textsc{replaced}, \textsc{removed}\},
    \label{eq:belief}
\end{array}
\end{equation}
which is updated recursively during revisits.

\subsection{System Pipeline Overview}
\label{sec:overview}
\begin{figure}
    \centering
    \includegraphics[width=1.0\linewidth]{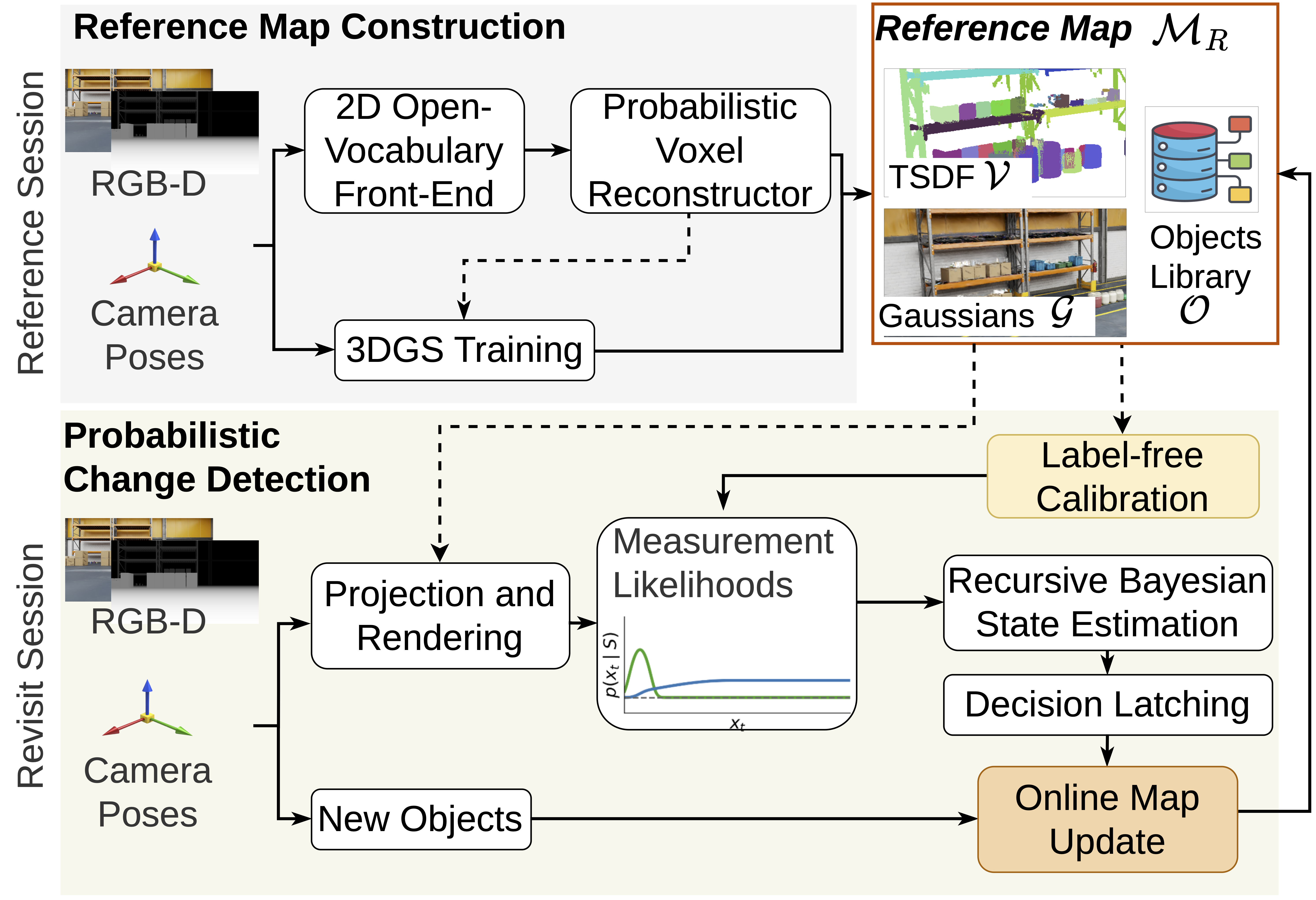}
    \caption{System Pipeline. The system first constructs an object-level reference map including a TSDF-based geometric map, and a 3DGS. During later revisits, the system infers each mapped object state by fusing geometry and appearance probabilistic evidence, and updates the map online.}
    \label{fig:pipeline}
\end{figure}
As shown in Fig.~\ref{fig:pipeline}, our system has two stages: building a reference map $\mathcal{M}_R$ in the initial traversal, and running change detection and maintenance in later revisits. The input is a sequence of aligned RGB-D frames $\mathcal{F} = \{F_t\}_{t=1\ldots T}$ with their camera-to-world poses $\mathcal{T} = \{T_t^{WC}\}_{t=1\ldots T}$. In the first stage, we incrementally build $\mathcal{M}_R$ from scratch, consisting of a TSDF voxel map $\mathcal{V}$ for geometry and a Gaussian Splatting model $\mathcal{G}$ for appearance. At the same time, we maintain an object library $\mathcal{O} = \{O_0,\ldots, O_n\}$. After construction, the reference recording is replayed on the reference map to calibrate the filter without labels.

In the second stage, we monitor the state of every mapped object against $\mathcal{M}_R$ online. Specifically, each object $O_i$ carries a three-state belief $b_i(S)$, Eq.~(\ref{eq:belief}), recursively updated as observations accumulate. Then, decisions are applied to the map during the revisit.
 
\subsection{Reference Map Construction}
\label{sec:refmap}

Building on OmniMap~\cite{deng2025omnimap}, we construct and maintain a geometry TSDF voxel map $\mathcal{V}$ and an appearance Gaussian Splatting model $\mathcal{G}$ in parallel. Objects are detected and segmented by an open-vocabulary front-end: we use DART \cite{turkcan2026detect}, built on SAM3 \cite{carion2026sam}, to obtain instance masks in real time, and then apply TAP~\cite{pan2024tokenize} to generate a refined segmentation mask with a description caption for each candidate. The caption is encoded into a semantic embedding by the SBERT encoder. We then project these 2D masks into 3D voxels using the depth. Following OmniMap, we adopt probabilistic voxel association. To be more specific, each voxel maintains several candidate instance IDs with their vote counts. Given a new mask, we compute a weighted sum of geometric similarity and semantic similarity for each candidate, and determine whether the mask is associated to an existing object or initializes a new one. After several confirmed observations, objects are added to the object library $\mathcal{O}$ and kept in the map.

During Gaussian Splatting training, we select keyframes by accumulated translation and rotation rather than at a fixed frame interval. On patrol-style trajectories, the robot speed varies substantially, so selecting keyframes by frame count makes the spatial density of the appearance reference vary with speed, resulting in uneven rendering quality.

The resulting reference map $\mathcal{M}_R = \{\mathcal{V}, \mathcal{G}, \mathcal{O}\}$ renders the reference scene from an arbitrary camera view, providing the geometric and appearance reference for each later revisit.

\subsection{Object State Representation}
Each mapped object $O_i$ keeps $\{V_i, \mathbf{f}_i, \hat a_i, \hat\ell_i, b_i\}$:
\begin{itemize}
  \item Voxel set $V_i$: the voxels the object occupies, labeled by majority vote during mapping.
  \item Semantic feature $\mathbf{f}_i$: the fused SBERT embedding.
  \item Appearance and geometric zero-points $\hat a_i, \hat\ell_i$: the residuals an unchanged object is expected to produce under the reference map, each with its posterior variance and per-object noise scale.
  \item Three-state belief $b_i(S)$, initialized to the prior $b_0$.
\end{itemize}

\section{Probabilistic Change Detection and Online Map Update}

\providecommand{\stateSame}{\textit{pr}\xspace}
\providecommand{\stateRp}{\textit{rp}\xspace}
\providecommand{\stateRm}{\textit{rm}\xspace}

\subsection{Observation Model}
\label{sec:obsmodel}

We use two channels of evidence to determine change: a geometry channel and an appearance channel. For each frame $F_t$ in a revisit session, we construct an observation for every mapped object visible in the current camera frustum,
\begin{equation}
    z_t = \{\Delta_t,\, x_t\},
    \label{eq:obs}
\end{equation}
where $\Delta_t$ denotes the geometric residual and $x_t$ denotes the appearance residual. We render each object's expectation and compare it with the current frame.

\textbf{Projection and rendering alignment.} Using current pose $T_t$, we project the voxels of all mapped objects into the current frame and keep, at each pixel, the object nearest to the camera. This generates a model depth image $d_t^{\mathrm{model}}$ and a per-pixel instance-label image. In parallel, we render the Gaussian field $\mathcal{G}$ from the same pose and get $\hat I_t$, the expected appearance of the reference scene from the current view. Both channels are aligned in the current view; then, each object's residuals are computed on its own pixel set.

\textbf{Residuals.} Residuals are computed over the core mask $\mathcal{C}_t$ of each object, the object's projected pixels after occlusion culling and a small erosion. The erosion drops the one- to two-pixel ring at silhouettes where the Gaussian field alpha-blends across depth discontinuities. The geometric residual is the median model--sensor depth discrepancy over $\mathcal{C}_t$, 
\begin{equation}
    \Delta_t = \operatorname*{median}_{m \in \mathcal{C}_t}\big(d_t^{\mathrm{model}}(m) - d_t^{\mathrm{sensor}}(m)\big).
    \label{eq:geo}
\end{equation}
When an object is removed, rays pass through its original location and terminate on the background, producing a large negative see-through residual. On the other hand, for both in-place \stateRp and \stateSame, $\Delta_t \approx 0$. This is the ambiguity the appearance channel is introduced to resolve. The appearance residual is the complement of a per-pixel multi-scale structural similarity map averaged over the core,
\begin{equation}
    x_t = 1 - \operatorname*{mean}_{m \in \mathcal{C}_t} M_t(m), \qquad M_t(m) = \prod_{l=1}^{3} S_l(m)^{w_l},
    \label{eq:app}
\end{equation}
where $S_l$ is the SSIM map between the rendering and the observation at pixel $m$ at scale $l$ with its weight $w_l$. $M_t(m)$ is a per-pixel similarity. Then the residual is the mean dissimilarity over the object's core area.


\subsection{Measurement Likelihoods and Label-Free Calibration}
\label{likelihood}
Given the state $S$, we assume the appearance channel and the geometry channel are conditionally independent. The measurement likelihood is
\begin{equation}
    p(z_t \mid S) = p(x_t \mid S)\, p(\Delta_t \mid S).
\end{equation}

\textbf{Appearance likelihood.} Exposure and rendering drift shift a whole frame at once, so we remove a per-frame offset $m_t$,
\begin{equation}
    \tilde x_t = x_t - m_t.
\end{equation} 
\begin{equation}
    m_t = \operatorname{median}_{i \in \mathcal{V}_t}
\left(x_t^{(i)} - \hat a_i\right),
\end{equation} 
$\mathcal{V}_t$ is the visible objects set in the current frame. Under \stateSame, the residual should fluctuate around the object's appearance zero-point: 
\begin{equation}
    p(\tilde x_t \mid \stateSame) = \mathcal{N}\big(\tilde x_t \mid \hat a_i,\ v_{a,i} + \sigma_{a,i}^2\big),
\end{equation}

where $ v_{a,i}$ is object's observation variance, and $\sigma_{a,i}^2$ is the posterior variance of its zero-point. 

Under a change, either \stateRp{} or \stateRm{}, the residual is expected to be lifted by a change magnitude $c$, 
which is constrained to be non-negative, since a change can only reduce similarity. Specifically, we place a truncated Gaussian prior on $c$, with mean $\mu_c$ and scale $\sigma_c$, and marginalize it out. The change likelihood has a closed form:
\begin{equation}
    p(\tilde x_t \mid \text{chg}) = \frac{\mathcal{N}\big(y \mid \mu_c,\ \sigma_c^2 + s^2\big)\ \Phi\big(\tilde m/\sqrt{\tilde v}\big)}{\Phi\big(\mu_c/\sigma_c\big)},
    \label{eq:app-chg}
\end{equation}
where $y = \tilde x_t - \hat a_i$, $s^2 = v_{a,i} + \sigma_{a,i}^2$, and $\tilde m, \tilde v$ are the posterior moments of $c$ given $y$, with $\tilde v = (1/\sigma_c^2 + 1/s^2)^{-1}$ and $\tilde m = \tilde v\,(\mu_c/\sigma_c^2 + y/s^2)$. Since the Gaussian prior decays quickly above $\mu_c$, we hold the likelihood at its peak beyond  $\mu_c$ to keep a larger difference from weakening the evidence.

\textbf{Geometry likelihood.} We consider \stateSame{} and \stateRp{} share the geometry likelihood, since the in-place replacement is unobservable to geometry.
\begin{equation}
    p(\Delta_t \mid \stateSame) = p(\Delta_t \mid \stateRp) = \mathcal{N}\big(\Delta_t \mid \hat\ell_i,\ \tau_i^2 + \sigma_{\ell,i}^2\big),
    \label{eq:geo-same}
\end{equation}
where $\hat\ell_i$ is the geometric zero-point, $\tau_i^2$ the object's depth observation variance, and $\sigma_{\ell,i}^2$ the posterior variance of the zero-point.
For the state \stateRm, following the Gaussian-uniform outlier treatment of POCD~\cite{qian2022pocd}, we consider a bounded uniform over the see-through range:
\begin{equation}
    p(\Delta_t \mid \stateRm) = \mathcal{U}\big(\Delta_t \mid [\delta_{\min}, \delta_{\max}]\big).
\end{equation}

\textbf{Robustness floor.} Single frames occasionally fail through unstable rendering, depth holes, or occlusion.
Rather than rejecting them by a manually chosen residual threshold, we apply $\varepsilon$-contamination to each likelihood, $p \mapsto (1-\varepsilon)p + \varphi$. We set the floor by capping how much one frame may contribute toward change at $\Lambda$ nats,

\begin{equation}
    \varphi = \frac{(1-\varepsilon)\, f_{\max}}{e^{\Lambda} - 1},
    \label{eq:floor}
\end{equation}
where $f_{\max}$ is the peak density of the change likelihood of the corresponding channel, applied per channel.

\textbf{Label-free calibration.} The per-object noise parameters and the decision threshold come from a change-free replay. The observations used to build the reference map are unchanged by construction. Therefore, after construction, we replay the reference session as if it were a revisit, with the map frozen, and compute the residuals for each object. These residuals contain no real change, and their spread is entirely rendering error and sensor noise. We split the frames into two interleaved halves, one for the per-object parameters and one for the decision threshold, so no frame serves both. For each object, the median of its corrected residuals is the raw estimate of its zero-point, and their spread gives the object's own observation variance $v_{a,i}$. Since the median is unreliable for objects seen only a few times, we combine it with a population prior through a conjugate update to get the zero-point $\hat{a}_i$, together with its posterior variance $\sigma_{a,i}^2$. The geometric quantities $\hat{\ell}_i$ and $\tau_i^2$ follow the same procedure. The change magnitude $\mu_c$ and $\sigma_c$ cannot be estimated from change-free data, so we fit them once as a prior and refine them causally during the revisit with a two-component mixture over the residuals seen so far, allowing the mean only to increase.

\subsection{Bayesian Recursion Update}
For each object, we run a Bayesian filter. The state has two parts, a discrete belief over the three states,  $b_t(S) = p(S \mid z_{1:t})$, and the Gaussian posteriors of two zero-points, $\mathcal{N}(\hat a_i, \sigma_{a,i}^2)$ and $\mathcal{N}(\hat\ell_i, \sigma_{\ell,i}^2)$ which capture the residual a persisted object is expected to produce. 
The belief starts at
\begin{equation}
b_0 = (1-\pi_0, \pi_0(1-\pi_{\mathrm{rm}}), \pi_0\pi_{\mathrm{rm}}),
\end{equation}
where $\pi_0$ is the prior probability of change, and
$\pi_{\mathrm{rm}}$ is the conditional prior probability that a change is a removal. The zero-points are initialized during calibration.

\textbf{Prediction.} 
Between observations, the belief reverts slightly toward the prior, so that evidence accumulated long ago does not dominate the current estimate,
\begin{equation}
    b^-_t = (1-\rho)\, b_{t-1} + \rho\, b_0.
    \label{eq:predict}
\end{equation}
The zero-points revert at the same rate toward their calibrated values, so that a zero-point driven off by earlier observations does not veto its own channel indefinitely.

\textbf{Update.} Each observation updates the belief with the two likelihoods,
\begin{equation}
    b_t(S) \propto b^-_t(S)\, \big[p(\tilde x_t \mid S)\big]^{w_t}\, \big[p(\Delta_t \mid S)\big]^{w_t},
    \label{eq:belief-update}
\end{equation}
tempered by a weight $w_t$ that accounts for correlated frames. Consecutive frames are far from independent. A robot that dwells on one viewing direction could turn a bad rendering into overwhelming evidence if all frames are used for updates. We bin the viewing directions around each object in azimuth and elevation, give the $n$-th observation in a bin the weight
\begin{equation}
    w_t = \sqrt{n}-\sqrt{n-1},
\end{equation}
so that the weights of the first $n$ observations in a bin telescope to exactly $\sqrt{n}$. In this way the discount is available online.

The zero-points are jointly updated with the belief, following POCD~\cite{qian2022pocd}. Whether the current frame should be fused into the zero-points depends on whether the object is still unchanged. Under \stateSame, the residual is a valid sample of the zero-point. Taking the appearance zero-point as an example, it is fused by a conjugate update, 

\begin{equation}
    \gamma^2 = \Big(\tfrac{1}{\sigma_{a,i}^2} + \tfrac{w_t}{v_{a,i}}\Big)^{-1},
    \qquad
    m = \gamma^2\Big(\tfrac{\hat a_i}{\sigma_{a,i}^2} + \tfrac{w_t \tilde x_t}{v_{a,i}}\Big).
    \label{eq:zp-branch}
\end{equation}
When the state is a change, the residual comes from the change rather than the zero-point and carries no information about it, so the branch remains at the prior, $m_{\mathrm{chg}} = \hat a_i$, $\gamma_{\mathrm{chg}}^2 = \sigma_{a,i}^2$. The exact posterior is a two-component Gaussian mixture of these two branches,
\begin{equation}
    p(a_i \mid z_{1:t}) = C_1\,\mathcal{N}(a_i \mid m, \gamma^2) + C_2\,\mathcal{N}(a_i \mid \hat a_i, \sigma_{a,i}^2),
    \label{eq:zp-mixture}
\end{equation}
where $C_1$ and $C_2$ are the normalized branch weights.
 
As the mixture is no longer a single Gaussian, we match its first two moments so the recursion can continue,
\begin{equation}
\begin{aligned}
    \hat a_i &\leftarrow C_1 m + C_2 \hat a_i, \\
    \sigma_{a,i}^2 &\leftarrow C_1 \gamma^2 + C_2\sigma_{a,i}^2 + C_1 C_2 (m - \hat a_i)^2.
\end{aligned}
\label{eq:zp-match}
\end{equation}
The geometric zero-point follows the same procedure.

\subsection{Decision and Online Map Update}
We define the posterior change log-odds as
\begin{equation}
    L_t = \log\frac{b_t(\stateRp) + b_t(\stateRm)}{b_t(\stateSame)}.
    \label{eq:logodds}
\end{equation}
A change is declared for an object as soon as $L_t$ crosses a threshold $\theta$ from the change-free calibration: we collect the per-object $\max_tL_t$ and take a high quantile at the acceptable false-positive rate. Then, the decision is latched, so \stateSame{} is ruled out and does not return, and the belief keeps updating between \stateRp{} and \stateRm{} over a short confirmation window. The change type follows the larger of $b_t(\stateRm)$ and $b_t(\stateRp)$. Once the type is fixed, the map is edited. The object's Gaussians are hidden, its TSDF region is reset, and a new instance is re-integrated from recently buffered frames. Any newly detected instance in the revisit session takes no further part in the belief update.

~\begin{table*}[!t]
\centering
\caption{Quantitative change detection results.}
\label{tab:change_results}
\renewcommand{\arraystretch}{1.08}
\setlength{\tabcolsep}{4pt}
\resizebox{\textwidth}{!}{%
\begin{tabular}{c l l c ccccc ccccc}
\toprule
& & & \multicolumn{1}{c}{\textbf{Persisted}} 
& \multicolumn{5}{c}{\textbf{Replaced}} 
& \multicolumn{5}{c}{\textbf{Removed}} \\
\cmidrule(lr){4-4} 
\cmidrule(lr){5-9} 
\cmidrule(lr){10-14}

\textbf{Scene} 
& \textbf{Method} 
& \textbf{Level} 
& \textbf{FP/N (\%)$\downarrow$} 
& \textbf{TP/P$^{*}$} 
& \textbf{FP/N (\%)$\downarrow$} 
& $\mathbf{P}\uparrow$ 
& $\mathbf{R}\uparrow$ 
& $\mathbf{F1}\uparrow$ 
& \textbf{TP/P$^{*}$} 
& \textbf{FP/N (\%)$\downarrow$} 
& $\mathbf{P}\uparrow$ 
& $\mathbf{R}\uparrow$ 
& $\mathbf{F1}\uparrow$ \\
\midrule
\multirow{6}{*}{\shortstack{Small\\Warehouse}}
& \multirow{2}{*}{Perceive-Semantix} 
& Voxel 
& \underline{8.2\%}
& 6242/37192 
& \underline{8.9\%}
& 0.11 
& 0.17 
& 0.13 
& 6692/28926 
& \underline{6.3\%}
& \underline{0.15}
& 0.23 
& \underline{0.18} \\

& 
& Object 
& \underline{33.8\%}
& 6/27 
& \underline{23.9\%}
& 0.19 
& 0.22 
& 0.20 
& 5/22 
& \underline{26.3\%}
& 0.14 
& 0.23 
& 0.18 \\

\cmidrule(lr){2-14}

& \multirow{2}{*}{DynamicGSG}
& Voxel
& 14.5\%
& \underline{15363/37192}
& 12.2\%
& \underline{0.17}
& \underline{0.41}
& \underline{0.25}
& \underline{9170/28926}
& 10.6\%
& 0.13
& \underline{0.32}
& \underline{0.18} \\

&
& Object
& 55.9\%
& \underline{15/27}
& 54.1\%
& \underline{0.20}
& \underline{0.56}
& \underline{0.30}
& \underline{14/22}
& 37.7\%
& \underline{0.25}
& \underline{0.64}
& \underline{0.35} \\

\cmidrule(lr){2-14}

& \multirow{2}{*}{\textbf{Ours}} 
& Voxel 
& \textbf{4.7\%} 
& \textbf{27385/37192} 
& \textbf{1.2\%} 
& \textbf{0.79} 
& \textbf{0.74} 
& \textbf{0.76} 
& \textbf{21073/28926} 
& \textbf{2.2\%} 
& \textbf{0.62} 
& \textbf{0.73} 
& \textbf{0.67} \\

& 
& Object 
& \textbf{10.3\%} 
& \textbf{27/27} 
& \textbf{9.2\%} 
& \textbf{0.73} 
& \textbf{1.00} 
& \textbf{0.84} 
& \textbf{22/22} 
& \textbf{15.8\%} 
& \textbf{0.55} 
& \textbf{1.00} 
& \textbf{0.71} \\

\midrule

\multirow{6}{*}{\shortstack{Large\\Warehouse}}
& \multirow{2}{*}{Perceive-Semantix} 
& Voxel 
& 19.6\% 
& 3649/22830 
& 20.3\% 
& 0.07 
& 0.16 
& 0.10 
& 3782/18063 
& 15.6\% 
& 0.09 
& 0.21 
& 0.13 \\

& 
& Object 
& 14.6\% 
& 18/82 
& 12.4\% 
& 0.08 
& 0.22 
& 0.12 
& 16/94 
& 13.1\% 
& 0.07 
& 0.17 
& 0.10 \\

\cmidrule(lr){2-14}
& \multirow{2}{*}{\textbf{Ours}} 
& Voxel 
& \textbf{3.2\%} 
& \textbf{18093/22830} 
& \textbf{4.9\%} 
& \textbf{0.62} 
& \textbf{0.79} 
& \textbf{0.70} 
& \textbf{14704/18063} 
& \textbf{10.4\%} 
& \textbf{0.38} 
& \textbf{0.81} 
& \textbf{0.51} \\

& 
& Object 
& \textbf{3.2\%} 
& \textbf{56/82} 
& \textbf{5.4\%} 
& \textbf{0.39} 
& \textbf{0.68} 
& \textbf{0.50} 
& \textbf{78/94} 
& \textbf{7.9\%} 
& \textbf{0.38} 
& \textbf{0.83} 
& \textbf{0.53} \\

\bottomrule
\end{tabular}%
}

\vspace{2pt}
\begin{minipage}{\textwidth}
\vspace{3pt}
\footnotesize
\footnotesize $^{*}$ $P$ denotes the number of evaluable GT positive objects under the common evaluation protocol. Best results are shown in bold, and second-best results are underlined. DynamicGSG could not be evaluated on the Large Warehouse due to GPU out-of-memory errors.
\end{minipage}
\vspace{-1mm}
\end{table*}
\section{Dataset}
\begin{figure}[H]
    \centering
    \includegraphics[width=0.98\linewidth]{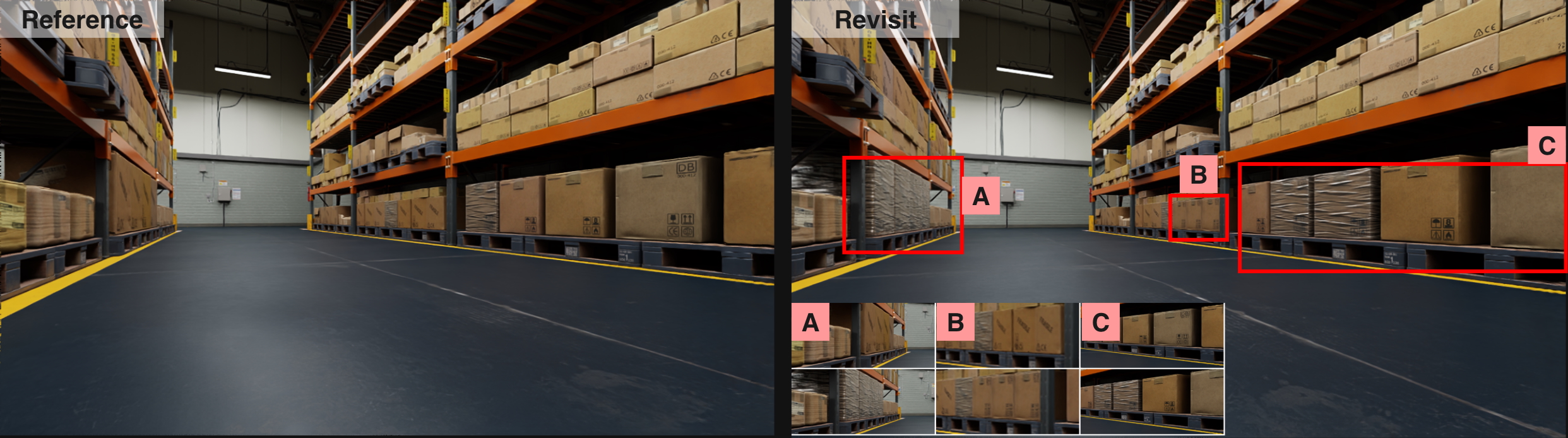}
    \caption{Three examples of in-place replacements (A--C) that preserve category and geometry but differ in wrapping, labels, or color. }
    \label{fig:replace}
\end{figure}

We build two simulated warehouse environments of different scales in NVIDIA Isaac Sim 5.1, as shown in Fig.~\ref{fig:replace}. A simulated Clearpath Husky drives through each environment repeatedly to mimic the revisits of an inspection robot. The scenes consist of common warehouse assets such as cardboard boxes, pallets, plastic drums, shelves, forklifts, and signs. Each traversal includes 1280$\times$720 aligned RGB-D images, instance-level segmentation labels, and camera poses.  Each object keeps a consistent global ID across multiple traversals, providing exact object association and change ground truth. For each warehouse, we record one reference sequence and multiple revisit sequences. More specifically, we create different changed revisit sessions by randomly removing objects, replacing objects, and changing object appearances or materials. 

\section{Experiments}
\label{sec:exp-warehouse}
\subsection{Experimental Setup}
\subsubsection{Baselines}

\textbf{Perceive-Semantix}~\cite{bogenberger2026glasses} extends POCD with an additional semantic-similarity likelihood. We take it as the representative of this family. 

\textbf{DynamicGSG}~\cite{ge2025dynamicgsg} detects changes based on 3DGS. It renders each mapped object from Gaussian map and deletes it once the single-frame SSIM against the observation falls below a fixed threshold. Since it runs out of GPU memory at full resolution, we run it at 640$\times$360 on the small warehouse and on TorWIC; the large-warehouse sequences are too long to run at any resolution.
\subsubsection{Evaluation Metrics}
We evaluate change-detection performance at two levels: voxel-level and object-level. At the voxel level, a ground-truth surface voxel counts as changed if the map segment covering it is declared changed by the method, and as a false positive if it belongs to an unchanged object. The voxel level does not require object-level instance matching and mainly focuses on spatial correctness. At the object level, the unit is the annotated physical object. Each detection is assigned to the ground-truth object containing the majority of its voxels, or treated as unmatched if it does not overlap any evaluable object. Both unmatched detections and detections assigned to unchanged objects are counted as false positives. 

 \begin{figure}[tbp]
    \centering
    \includegraphics[width=0.98\linewidth]{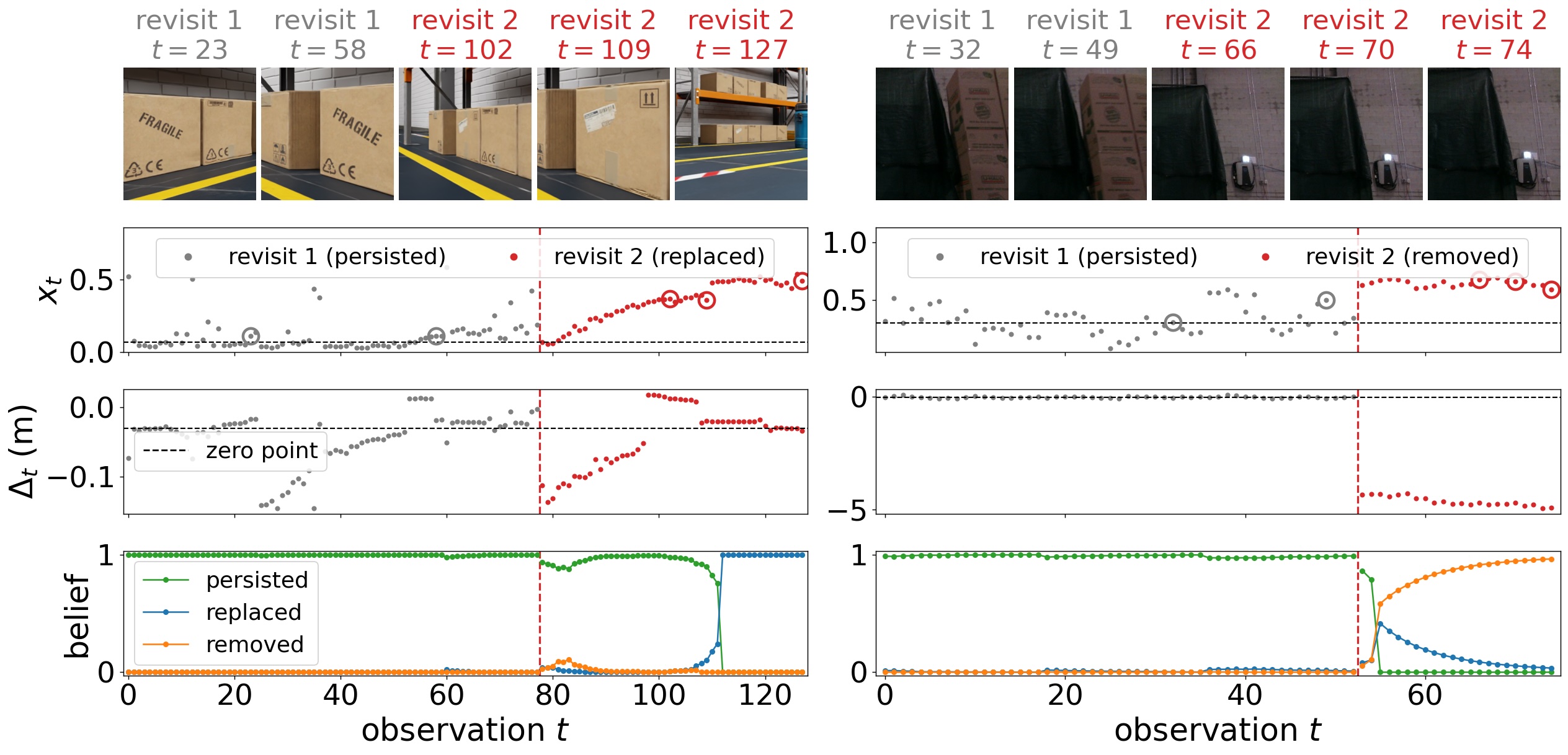}
    \vspace{-2mm}
    \caption{\textbf{Evidence and object-state belief.} An object is observed unchanged in revisit~1 (gray) and replaced by a same-sized box in revisit~2 (red); top frames correspond to the circled observations. \textbf{Left.} In-place replacement. The geometric measurement $\Delta_t$ is indistinguishable between the two revisits, whereas the appearance measurement $x_t$ clearly separates them. The recursive belief ignores isolated outliers and converges
  to \textsc{replaced} once appearance evidence accumulates. \textbf{Right.} Real-world removal case.}
    \label{fig:belief}
\end{figure}

\subsection{Isaac Sim Warehouse Dataset}
Each warehouse provides four sequences. A reference traversal is denoted $t_0$, two revisits with replacement and removal  are denoted $t_{\mathrm{rp}}$ and $t_{\mathrm{rm}}$, and a static revisit $t_{\mathrm{static}}$, in which nothing changes, is used to measure false detections. Table~\ref{tab:change_results} reports the results under the two-level protocol.

We first compare all methods on $t_{\mathrm{static}}$. Since nothing changes, every detection is a false positive. The object-level false-positive rate of Perceive-Semantix is 14.6\% on the large warehouse and 33.8\% on the small one, and DynamicGSG flags 55.9\% of the unchanged objects on the small warehouse, since a single-frame SSIM drop is enough for it to delete an object. Ours is 3.2\% and 10.3\%. For our method this also shows that the threshold $\theta$, calibrated by replaying $t_0$, transfers to the different trajectory of $t_{\mathrm{static}}$. Fig.~\ref{fig:belief} illustrates how evidence accumulates for a box replaced by one of identical size but with a different label.

On changed revisits our method also shows high recall in all cases. It detects all changed objects in the small warehouse and more than 68\% of them in the large warehouse, while Perceive-Semantix's recall does not exceed 0.23 in any case, and DynamicGSG reaches at most 0.64 while flagging more than half of the unchanged objects. In particular, the gap is widest on replacement, the case our appearance model is designed for. For Perceive-Semantix, replacement is harder than removal, with a voxel-level F1 of 0.13 on replacement against 0.18 on removal in the small warehouse. 
For our method the two cases are comparable, with 0.76 against 0.67.
\begin{figure}[H]
  \centering
  \includegraphics[width=0.95\linewidth]{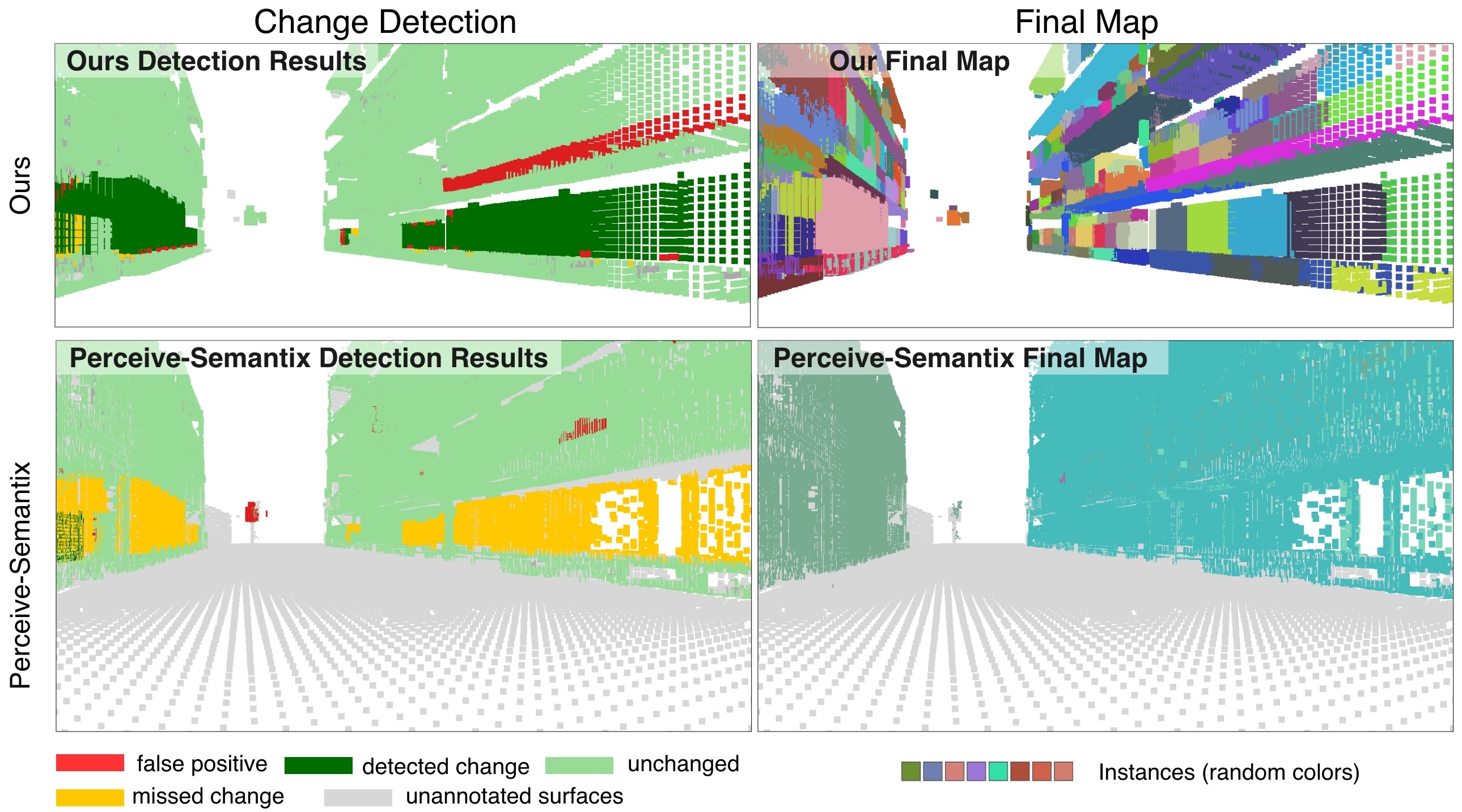}
  \vspace{-2mm}
  \caption{\textbf{Fine-grained change detection and map maintenance in large warehouse.} Our method maintains a clean instance-level map, whereas Perceive-Semantix misses most changes and merges adjacent boxes into large semantic segments.}
  \label{fig:qualitative}
\end{figure}
Perceive-Semantix suffers mainly from association. Warehouse assets are packed tightly, and its association merges new detections into old objects 591 times on the large warehouse, which also explains its high false-positive rate on $t_{\mathrm{static}}$, as shown in Fig.~\ref{fig:qualitative}. Our method has a failure mode of its own. On the large-warehouse removal revisit the object-level F1 drops to 0.53, and in both scenes the false-positive rate on removal is higher than on replacement. We attribute this to online map maintenance. Once an object is removed, it exposes surfaces of neighbouring objects that were never observed during reference mapping, whose rendered appearance then mismatches the observation and leads them to be flagged as changed.

Table~\ref{tab:state_cls} shows that the estimator also assigns the right kind of change to nearly all detected objects, with the most common confusion being some removed objects in the large warehouse labelled as replaced.
~
\begin{table}[H]
\centering
\caption{State Classification of Detected Changed Objects}
\label{tab:state_cls}
\renewcommand{\arraystretch}{1.1}
\setlength{\tabcolsep}{3pt}
\resizebox{\columnwidth}{!}{%
\begin{tabular}{lcc cc}
\toprule
& \multicolumn{2}{c}{\textbf{GT Replaced}}
& \multicolumn{2}{c}{\textbf{GT Removed}} \\
\cmidrule(lr){2-3}\cmidrule(lr){4-5}
\textbf{Scene}

& \textbf{Pred. Rep.}
& \textbf{Pred. Rem.}

& \textbf{Pred. Rep.}
& \textbf{Pred. Rem.} \\
\midrule
Small Warehouse
 & 25 & 2
 & 1 & 21 \\

Large Warehouse
& 52 & 4
 & 11 & 67 \\
\bottomrule
\end{tabular}%
}

\end{table}
\subsection{TorWIC Real-World Warehouse Dataset}
We evaluate our method on the real-world TorWIC dataset~\cite{qian2022pocd}, which was collected with a RealSense D435i RGB-D camera on a mobile robot, with poses from LiDAR-based SLAM. We evaluate four box- and fence-removal scenarios, \texttt{2-1} to \texttt{2-4}. We evaluate all methods at the voxel level, as TorWIC provides only placement layouts to describe the changes, without object-level annotation. Since these sequences contain removals only, the evaluation reduces to a binary decision, changed or unchanged.
\begin{table}[H]
\centering
\caption{Voxel-level removal detection results on TorWIC}
\label{tab:torwic}
\renewcommand{\arraystretch}{1.08}
\setlength{\tabcolsep}{3.5pt}
\resizebox{\columnwidth}{!}{%
\begin{tabular}{llccccc}
\toprule
\textbf{Scene}
& \textbf{Method}
& \textbf{TP/P}
& \textbf{FP/N (\%)$\downarrow$}
& $\mathbf{P}\uparrow$
& $\mathbf{R}\uparrow$
& $\mathbf{F1}\uparrow$ \\
\midrule

\multirow{3}{*}{2-1}
& Perceive-Semantix
& 9,493/11,378
& \underline{20.1}
& \underline{0.37}
& 0.83
& \underline{0.51} \\
& DynamicGSG
& \textbf{9,973/11,378}
& 37.5
& 0.25
& \textbf{0.88}
& 0.39 \\
& Ours
& \underline{9,704/11,378}
& \textbf{6.0}
& \textbf{0.67}
& \underline{0.85}
& \textbf{0.75} \\
\midrule

\multirow{3}{*}{2-2}
& Perceive-Semantix
& 11,691/18,991
& \underline{16.5}
& \underline{0.52}
& 0.62
& \underline{0.56} \\
& DynamicGSG
& \underline{12,415/18,991}
& 23.4
& 0.45
& \underline{0.65}
& 0.53 \\
& Ours
& \textbf{13,716/18,991}
& \textbf{3.1}
& \textbf{0.87}
& \textbf{0.72}
& \textbf{0.79} \\
\midrule

\multirow{3}{*}{2-3}
& Perceive-Semantix
& 22,102/38,162
& 1.1
& \underline{0.98}
& 0.58
& 0.73 \\
& DynamicGSG
& \underline{25,268/38,162}
& \underline{1.0}
& \underline{0.98}
& \underline{0.66}
& \underline{0.79} \\
& Ours
& \textbf{28,406/38,162}
& \textbf{0.7}
& \textbf{0.99}
& \textbf{0.74}
& \textbf{0.85} \\
\midrule

\multirow{3}{*}{2-4}
& Perceive-Semantix
& 170/42,042
& \underline{58.6}
& 0.01
& 0.00
& 0.01 \\
& DynamicGSG
& \underline{3,794/42,042}
& 64.4
& \underline{0.12}
& \underline{0.09}
& \underline{0.10} \\
& Ours
& \textbf{31,012/42,042}
& \textbf{7.5}
& \textbf{0.91}
& \textbf{0.74}
& \textbf{0.81} \\

\bottomrule
\end{tabular}%
}
\begin{flushleft}
\footnotesize
Best results are shown in bold, and second-best results are underlined.
\end{flushleft}
\end{table}

\begin{figure}[!t]
    \centering
    \includegraphics[width=0.9\linewidth]{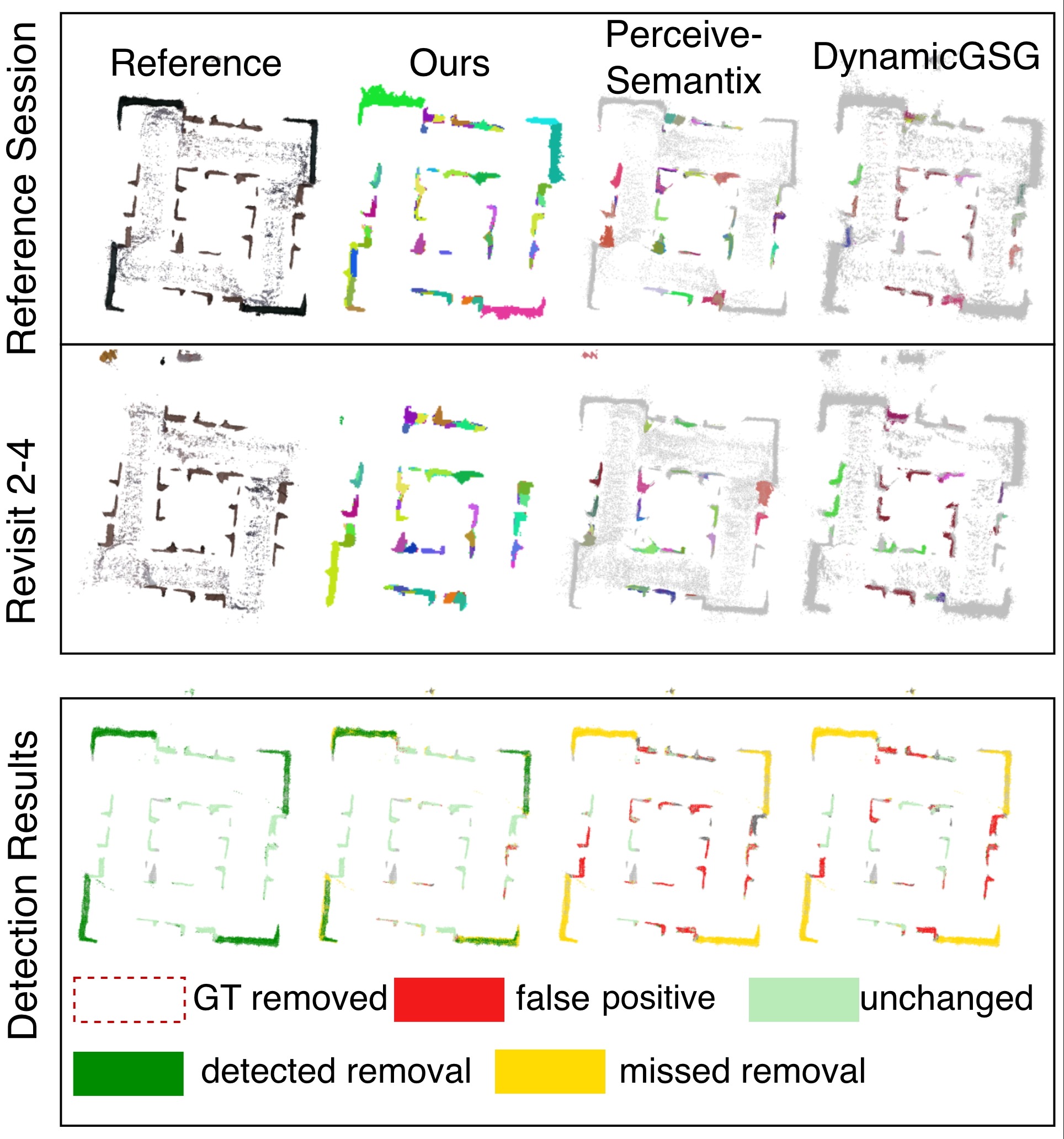}
    \vspace{-2mm}
    \caption{Results on TorWIC \texttt{2-4}. \textbf{Top}: instance-level maps colored by object ID, so a color preserved from the reference means the object kept its identity. Ours keeps most identities across the revisit, while Perceive-Semantix reconstructs the scene well but under new IDs. \textbf{Bottom}: detections against the layout ground truth, which show the same pattern.}
    \label{fig:torwic}
\end{figure}

As shown in Table~\ref{tab:torwic}, Perceive-Semantix performs better here than in simulation, reaching a recall of 0.83 on \texttt{2-1}. This recall comes with a false-positive rate of 20.1\%, and it tends to declare an object missing and rebuild a new one in its place. DynamicGSG reaches the highest recall on \texttt{2-1}, but at a false-positive rate of 37.5\%, since a single-frame SSIM drop is enough for it to delete an object. As Fig.~\ref{fig:torwic} shows, in \texttt{2-4}, where only the fence moved, both baselines fail. Perceive-Semantix treats the fence as background and never updates it, and DynamicGSG is at 64.4\% false positives.

For our method, migrating from the simulated dataset to TorWIC requires no manual parameter tuning, as we discussed in Sec.~\ref{likelihood}. The only change is the estimation of $m_t$. Only a few mapped objects are visible per frame, so the object-level median is unreliable and we estimate $m_t$ from background pixels such as the floor instead. Our overall recall is lower than on the simulated warehouses because SLAM drift and depth noise broaden the residual distribution of unchanged objects and raise the calibrated threshold.

\subsection{Ablation Studies}
We ablate the system on both warehouses at the voxel level. All variants run detection only with the map fixed, which isolates each component from the side effects of map edits. Table~\ref{tab:closedloop} shows this to be a fair proxy, as the fixed map matches the closed loop on replacement in both scenes and exceeds it on removal by 0.07 F1 in the large warehouse and 0.05 in the small one.

\begin{figure}[H]
    \centering
    \includegraphics[width=0.9\linewidth]{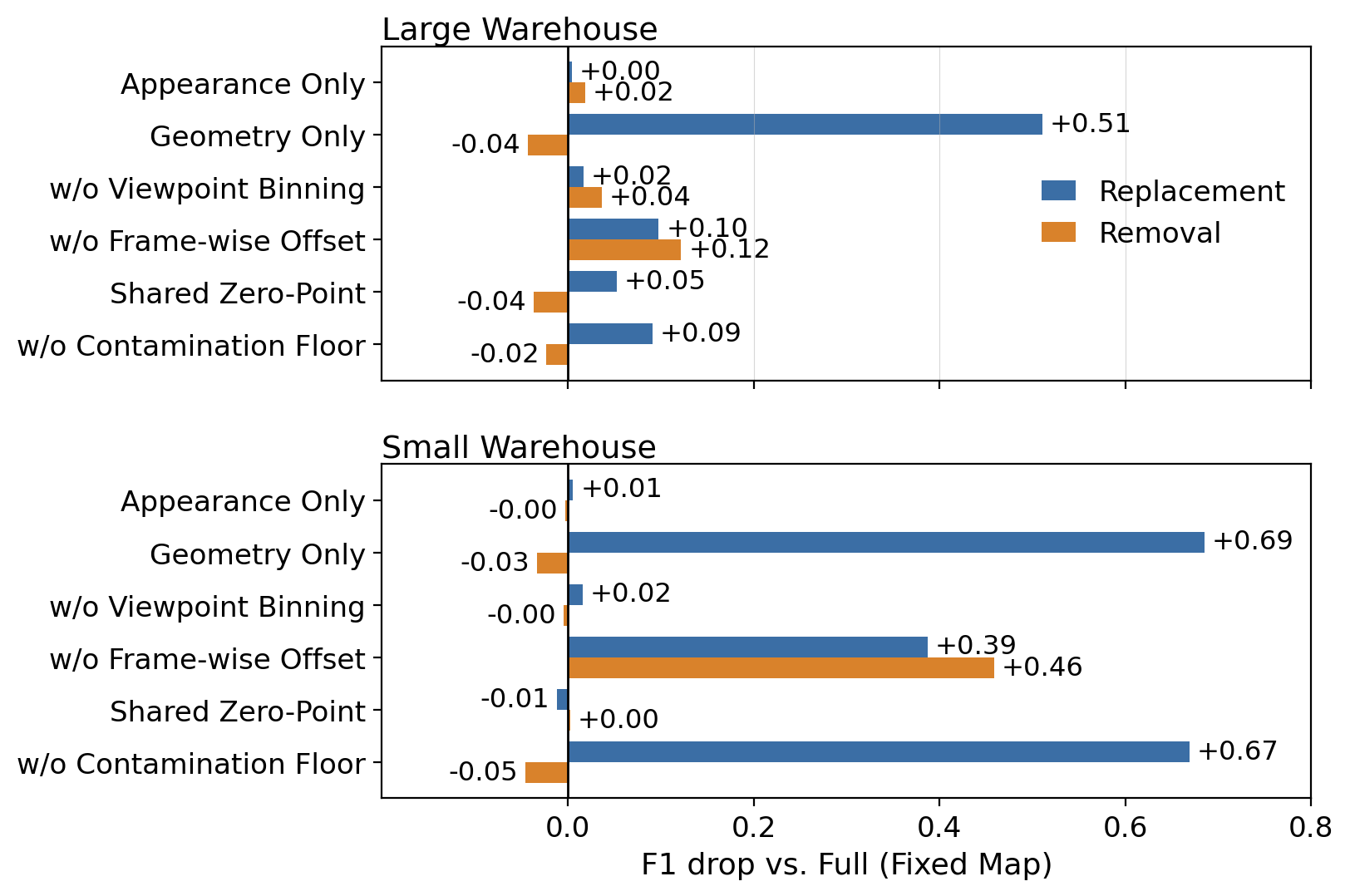}
    \vspace{-2mm}
    \caption{Ablation on the warehouse dataset. Bars show the F1 drop of each variant relative to the full model with the map fixed. Positive values mean the component helps while negative mean removing it improves that revisit.}
    \label{fig:ablation}
\end{figure}

As shown in Fig.~\ref{fig:ablation}, the two channels divide the work as designed. Geometry alone detects removal well with the fewest false positives but fails on replacement. For in-place replacement, the occupied volume does not change and geometry is unobservable. Appearance alone matches the full model on replacement at a higher false-positive rate that geometry then reduces. Removing the frame-wise offset $m_t$ lowers F1 in all four settings, by 0.10 and 0.12 in the large warehouse and 0.39 and 0.46 in the small one. The contamination floor mainly affects replacement. Without the per-frame cap, a few static objects accumulate extreme log-odds during replay, raising the calibrated threshold. The per-object zero-point and viewpoint binning address imperfect 3DGS rendering on real data, which is largely absent in simulation; removing either changes F1 by at most 0.05.

\begin{table}[tbp]
\centering
\caption{Effect of online map editing on voxel-level performance}
\label{tab:closedloop}
\renewcommand{\arraystretch}{1.05}
\setlength{\tabcolsep}{3.2pt}
\scriptsize
\resizebox{\columnwidth}{!}{%
\begin{tabular}{ll ccc ccc}
\toprule
& & \multicolumn{3}{c}{\textbf{Replaced}}
& \multicolumn{3}{c}{\textbf{Removed}} \\
\cmidrule(lr){3-5}
\cmidrule(lr){6-8}
\textbf{Scene} & \textbf{Variant}
& \textbf{FP/N (\%)$\downarrow$}
& $\mathbf{F1}\uparrow$
& \textbf{R@5\%$\uparrow$}
& \textbf{FP/N (\%)$\downarrow$}
& $\mathbf{F1}\uparrow$
& \textbf{R@5\%$\uparrow$} \\
\midrule
\multirow{2}{*}{Large}
& Full (closed loop) & 4.9 & 0.70 & 0.79 & 10.4 & 0.51 & 0.44 \\
& Full (map fixed)   & 4.3 & 0.72 & 0.80 &  7.6 & 0.58 & 0.61 \\
\midrule
\midrule
\multirow{2}{*}{Small}
& Full (closed loop) & 1.2 & 0.76 & 0.75 &  2.2 & 0.67 & 0.73 \\
& Full (map fixed)   & 1.0 & 0.77 & 0.76 &  1.4 & 0.72 & 0.73 \\
\bottomrule
\end{tabular}%
}
\vspace{4pt}
\begin{minipage}{\columnwidth}
\vspace{3pt}
\footnotesize
R@5\% denotes recall at a 5\% false-positive rate.
\end{minipage}
\vspace{-4mm}
\end{table}

\subsection{Computation Time}
Our system runs on an RTX 5090. Reference mapping is performed once per map, taking 1.8\,s per frame on the 1280$\times$720 warehouse sequences and 0.6\,s on the 640$\times$360 TorWIC sequences plus a calibration replay. Each revisit costs 0.73\,s per frame on the warehouses and 0.27\,s on TorWIC, dominated by segmentation, rendering, and TSDF integration, while the change filter itself takes 0.2\,ms per object observation. Perceive-Semantix costs 1.2\,s and 0.2\,s, and DynamicGSG 3.3\,s and 2.3\,s at 640$\times$360.

\section{Conclusions}
We presented FOCUS, an uncertainty-aware framework for object-level change detection and map maintenance in semi-static environments. FOCUS combines geometric consistency with view-conditioned appearance from a 3D Gaussian map and recursively estimates whether each object is \textsc{persisted}, \textsc{replaced}, or \textsc{removed}. By modeling appearance uncertainty and correlated observations, the method detects ambiguous in-place replacements that are difficult to identify from geometry or coarse semantics alone. Experiments on simulated warehouses and the real-world TorWIC dataset demonstrate improved change detection and map maintenance, highlighting the importance of uncertainty-aware appearance reasoning for persistent object-centric memory.

\bibliographystyle{IEEEtran}
\bibliography{bibliography/papers}

\clearpage
\onecolumn
\appendices

\end{document}